\documentclass[conference,a4paper]{IEEEtran}

\ifCLASSINFOpdf
   \usepackage[pdftex]{graphicx}
   \usepackage{epstopdf}
   \usepackage{afterpage}
\else

\fi

\usepackage{color}

\newcommand{\diego}[2]{\textcolor [rgb]{1,0.2,0}{#2}}
\newcommand{\mitch}[2]{\textcolor [rgb]{0.2,0.2,0.6}{#2}}

\begin{document}

\title{Learning  rotation invariant convolutional filters for texture classification}

\author{\IEEEauthorblockN{Diego Marcos, Michele Volpi and Devis Tuia}

\IEEEauthorblockA{MultiModal Remote Sensing\\
Department of Geography, University of Zurich, Switzerland
\\\texttt{$\{$diego.marcos,michele.volpi,devis.tuia$\}$@geo.uzh.ch}}}

\maketitle

\begin{abstract}
We present a method for learning discriminative\diego{steerable}{} filters using a shallow Convolutional Neural Network (CNN). We encode rotation invariance directly in the model by tying the weights of groups of filters to several rotated versions of the canonical filter in the group. These filters can be used to extract rotation invariant features well-suited for image classification. We test this learning procedure on a texture classification benchmark, where the orientations of the training images differ from those of the test images. We obtain results comparable to the state-of-the-art. Compared to standard shallow CNNs, the proposed method obtains higher classification performance while reducing by an order of magnitude the number of parameters to be learned.
\end{abstract}

\IEEEpeerreviewmaketitle

\vspace{2mm}
\section{Introduction}
Texture often contains useful \diego{discriminant}{} information about the materials composing the objects present in an image. It is therefore of great help for tasks such as image segmentation and classification~\cite{malik2001contour}. When considering texture, it is very useful to use appearance descriptors that are invariant to specific transformations, such as rotation, translation and scaling, since, in most cases, a material must be identified as such irrespectively of its relative position with respect to the camera. Rotation invariance is of particular interest, given that many computer vision problems consider images that are arbitrarily oriented. Some examples are remote sensing and microscopy images, where the interpretation of the images should not be affected by any global rotation in the spatial domain.

\subsection{Rotation invariance in texture classification}

Given a function $f$ and a transformation $\rho$, we consider $f$ as invariant with respect of $\rho$ if $f(x) = f(\rho(x))$ and equivariant if $\rho(f(x)) = f(\rho(x))$.
One approach to achieve invariance to rotation in anisotropic textures is to find the main axis of a query texture patch and rotate it until this axis is aligned with those in the training set~\cite{jafari2005radon}. Although intuitive, this approach has the disadvantage of requiring the main axis of the texture to be homogeneous within the whole patch and assuming that this axis can be retrieved efficiently and accurately.
Other common strategies define features that are rotation invariant or equivariant, \textit{i.e} features whose output values are not affected by rotations of the input image or whose outputs are rotated in the same way as the input image. Some examples are the rotation invariant Local Binary Patterns~\cite{ojala2002multiresolution}, spiral resampling~\cite{wu1996rotation}, steerable pyramid filters~\cite{greenspan1994rotation} and steerable wavelets~\cite{do2002rotation,depeursinge2014rotation}. 
Steerable filters~\cite{freeman1991design} have been introduced to analyze directional properties of images based on edge and orientation detection. They were designed such that they could be rotated to an arbitrary orientation without the need of interpolation. 
We propose to learn rotatable filters in a data-driven fashion. These filters are not strictly steerable in the sense of~\cite{freeman1991design}, since we do perform interpolation. Instead, we define as rotatable a filter that has been designed to be applied at different orientations and where all orientations are explicitly considered to have the same meaning\diego{(often referred to as weight tying)}{} and to contribute equivalently to the output features, what guarantees rotation equivariance.

\subsection{Rotation invariance in convolutional neural networks}

\begin{figure*}[t]
	\centering
	\includegraphics[width=.9\linewidth]{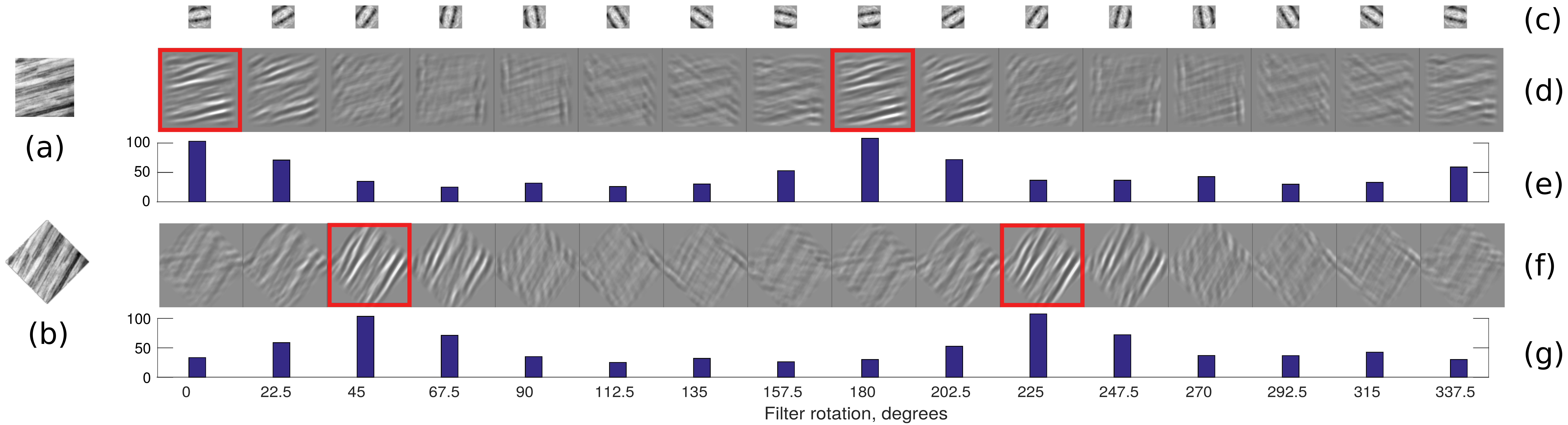}
	\caption{Given an image (a) and a version rotated by $45^{\circ}$ (b), the responses to multiple rotated versions of the same filter (c) are shown in (d) and (f) respectively. The maximally activated responses are marked with a red box. The bar plots (e) and (g) represent the maximal value over each feature map. Note how (g) is a shifted version of (e) and their maximal values are very similar. These maximal values are thus invariant to rotations of the input images (a) and (b).}
	\label{fig:concept}
\end{figure*}

In this work, we consider a shallow convolutional neural network (CNN,~\cite{lecun1998gradient}) to learn the set of rotatable filters. CNNs are well suited to tackle problems involving images through the use of learned convolutions and pooling operations. The convolutional layers enforce translation equivariance and encode a prior (shared weights through the use of convolutions) that reduces the total number of parameters to be learned by the neural network. Every activation value is computed by a linear combination of the image values in a fixed neighborhood, plus an independent bias term, and can therefore be expressed in the form of a convolution. This has the additional advantage of enabling the use of existing efficient algorithms to compute the convolutions, such as those based on the Fast Fourier Transform (FFT). The pooling operations of the network achieve local translation invariance by pooling values in small windows into a single activation. Depending on the architecture, CNNs can achieve a high level of invariance to local translations and deformations, which contributes to a certain degree to rotation invariance, although not explicitly encoded in the model.

Within the CNN framework, two approaches can be adopted to encode rotation invariance: 1) applying rotations to the feature maps (or the input image) or 2) applying rotations to the convolution filters.
The first category includes the common practice of augmenting the training set with many rotated versions of the training images, which allows the model to learn rotation invariance without explicitly enforcing it~\cite{simard2003best}. The disadvantage of this approach is that this results in a model requiring many redundant parameters, which are learned independently for each rotation angle. This is because nothing prevents the model to learn filters depicting the same patterns in different orientations. An alternative to this strategy is to simultaneously feed several rotated and flipped versions of the input image to the same deep CNN~\cite{fasel2006rotation,dieleman2015rotation}; this has a similar effect to data augmentation, as it promotes parameter sharing (the same filters are applied to the different rotated versions of the images), but again without explicitly enforcing invariance. This idea was extended in~\cite{dieleman2016exploiting} within several new neural network blocks, including a pooling operation over the rotations, aimed at reducing the number of parameters and making the rotation invariance more explicit.
Another view on this problem has been provided in~\cite{laptev2016ti}, where many rotations are applied to the input and the maximum activation over these is taken as the output of a fully connected layer in a CNN.

The second category applies rotations to the learnable filters. In~\cite{schmidt2012learning,kivinen2011transformation,sohn2012learning}, the authors propose to learn unsupervised rotation equivariant features by tying the weights of a Convolutional Restricted Boltzmann Machine (C-RBM). The authors of~\cite{teney2016learning} propose to tie the weights within several layers of a deep CNN by splitting the filters into orientation groups. Thereby, the model achieves rotation \textit{covariance}, \textit{i.e.} applying a rotation to the input image results in a shift of the output across the features. Therefore, it only needs to learn a single canonical filter per group. In~\cite{gens2014deep}, it is proposed to use pooling along several symmetry groups in order to achieve invariance to such symmetries. In~\cite{wu2015flip}, authors applied this concept to the rotation group in the upper layers of a deep CNN in order to obtain invariance to global rotations of the input image.

\subsection{Contribution}

We propose to learn tied, rotatable filters as a part of standard CNN training. Our contribution differs from those reviewed above, since the architecture of our shallow CNN provides explicit full invariance to the rotation of the input images. Moreover, such invariance is not passively encoded by simply constraining the family of filters that can be learned (\textit{e.g.} by enforcing filters with radial symmetry), but it allows for full discriminative learning of weights as in standard CNN convolutional layers. We study how tying the weights of several rotated versions of the filters in the first layer of a shallow CNN can be used to learn a set of rotatable filters for the extraction of rotation invariant features for texture classification. We pay particular attention to the filters learned, as well as the generalization ability of the model in small sample scenarios, i.e. scenario where the amount of training data is limited. This is of particular importance for many applications and even more when using CNN models, which are well known to be dependent on the availability of large collections of labeled data to perform well in practice.

\section{Learning rotatable filterbanks with CNNs}\label{sec:CNN}

We use a shallow CNN consisting of only one convolutional layer, several pooling blocks and a fully connected layer followed by a softmax loss for classification. The filters learned by the first convolutional layer can subsequently be used as a standard filterbank for texture classification independently from the rest of the network (\emph{i.e.} we can train any classifier using the CNN features as inputs), or the CNN can be used directly as the classifier (the output of the softmax\mitch{loss}{}). In the proposed pipeline, described in detail in Section~ \ref{sec:sota}, the shallow CNN is mainly used to learn features, rather than to provide the final classification scores for test samples.

Equivariance to rotations is achieved in the convolutional layer by tying the weights of the filters within the same rotation group, such that each filter in the group corresponds to a rotated version of the first filter in the group (the \textit{canonical} filter). Given a canonical filter $h_0^i$ in the $i^{th}$ rotation group, the other filters in the group are computed as:
\begin{equation}
	h_{\alpha}^i = \texttt{rotate}(h_0^i,\alpha) \quad \forall \alpha \in [1\dots R-1]\frac{2\pi}{R}
\end{equation}
where the \texttt{rotate} operator implements standard image rotation by bicubic interpolation and $R$ is the total number of angles taken into account in each rotation group. It is important to ignore the effect of the pixels in $h_0^i$ that are located outside the filter in some of the rotations. We therefore only use the pixels within a circle circumscribed in the square filter.

After this convolutional layer, we apply an orientation max-pooling operation, {\it i.e.} a max-pooling operation applied in the orientation dimension and returning the maximal value across the activations of the same rotation group. It ensures invariance to local rotations and therefore equivariance to global rotations of the input image. In Fig.~\ref{fig:concept} we show how this invariance is obtained. We train the CNN by standard backpropagation. In the backward pass, gradients are passed through the angle generating the largest activation, very similarly to how standard max-pooling behaves. The index of the angle activating the orientation max-pooling is recorded during the forward pass and used in the backward pass to update the weights of the canonical filter.
These two blocks, the rotatable convolutional layer and orientation max-pooling, have been implemented as modules in the MatConvNet~\cite{vedaldi2015matconvnet} CNN library for Matlab\footnote{www.vlfeat.org/matconvnet/}. 

\begin{figure}[t]
	\centering
	\includegraphics[width=\linewidth]{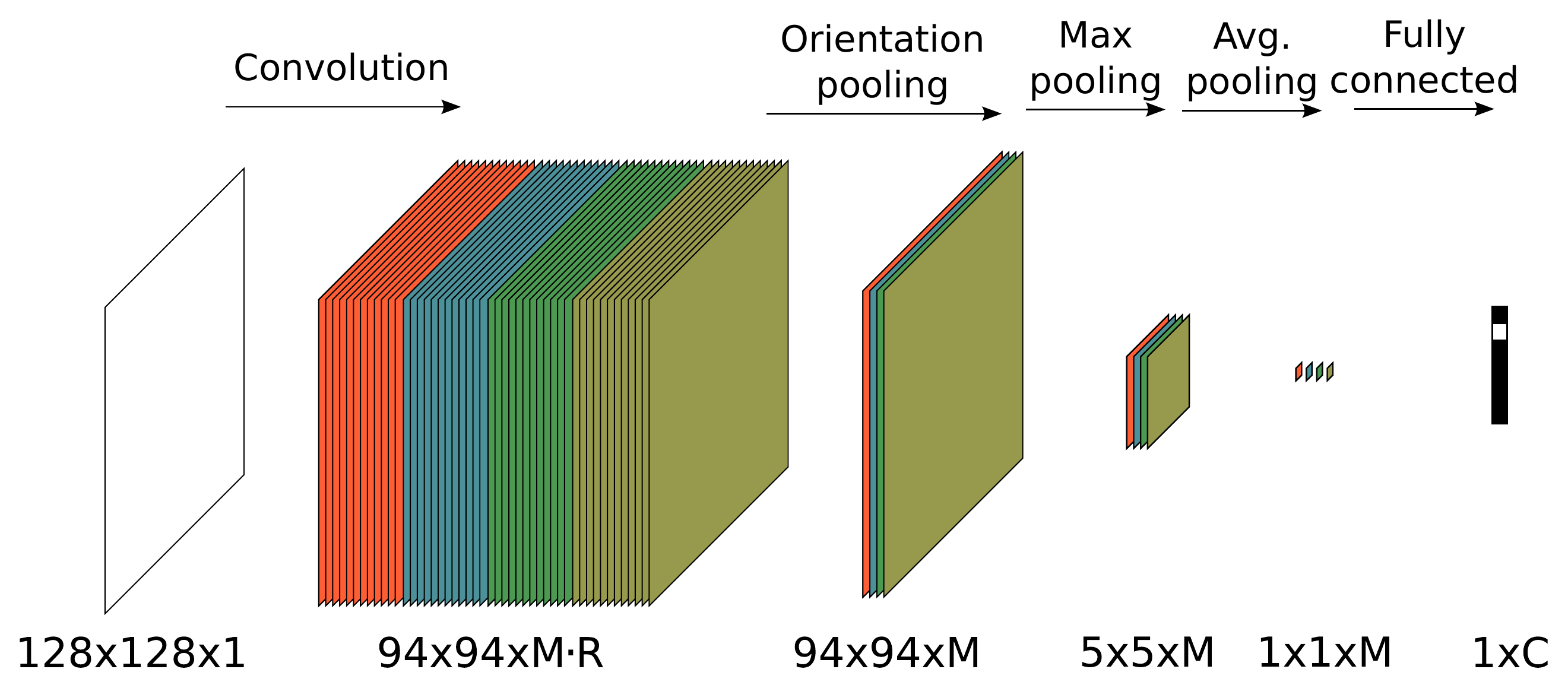}
	\caption{Architecture of the CNN used to learn the rotatable filters. }
	\label{fig:diag}
\end{figure}

We apply rectified linear units (ReLU) to the output of the orientation max-pooling.
Then, we add two spatial pooling blocks to obtain full rotation invariance: a max-pooling followed by an average-pooling. This last step allows the model to learn from several locations in each training image: one for each max-pooling window. Fig.~\ref{fig:diag} shows a diagram with the proposed network architecture. $M$ is the total number of rotation groups (\textit{i.e.} the total number of unique canonical filters) and $R$ is the number of discrete orientations considered within each group. $C$ is the number of classes in the classification problem.

The CNN parameters are learned using stochastic gradient descent (SGD) with momentum (as implemented in MatConvNet~\cite{vedaldi2015matconvnet}) and dropout~\cite{hinton2012improving}. The latter consists of the omission of a given percentage of randomly selected filters from the CNN at each weight update iteration. Dropout helps to prevent overfitting, making the learned filters more general. In addition, we use weight decay on the convolution filters (but not on the biases). Weight decay is a regularizer that favors parameters of small magnitude, thus eliminating spurious large parameters and helping convergence.

\section{Experimental setup}

\subsection{Dataset}

\begin{figure}[h]
	\centering
	\includegraphics[width=.95\linewidth]{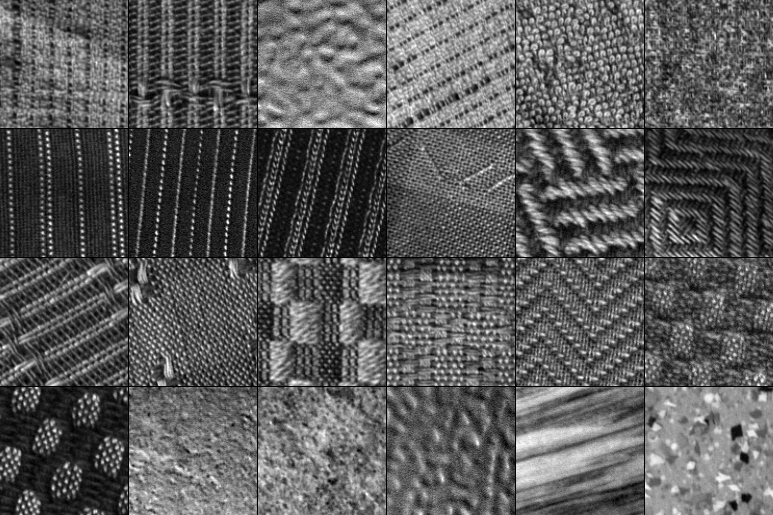}
	\caption{One image from the training set for each of the 24 texture classes in the Outex dataset acquired using illumination \textit{inca}.}
	\label{fig:data}
\end{figure}

We test the proposed method in 3 Outex\footnote{http://www.outex.oulu.fi} texture classification benchmarks aimed at testing rotation invariant methods. Outex\_TC\_00010 consists of a training set of grayscale texture photographs taken at one particular orientation and a test set taken at 8 different orientations not included in the training set. It contains 24 texture classes and 20 samples per class and orientation of size $128\times 128$ pixels each (Fig.~\ref{fig:data}). All the samples are acquired under an illumination called \emph{inca}. Outex\_TC\_00012 is similar, but more challenging because different illumination conditions not present in the training set have been included in the test set. For this dataset, two settings are generally considered: problem 000, acquired under illumination conditions labeled \textit{tl84} and problem 001, acquired under illumination \textit{horizon}. In both cases, the training set has been acquired using illumination conditions \textit{inca}. For all the experiments we randomly selected a $2\%$ of the \textit{inca} dataset as a holdout validation set for model selection.

\subsection{Texture classification pipeline}
\label{sec:sota}

In the first experiment we compare the proposed strategy with state-of-the-art rotation invariant classification methods. As mentioned above, we used the CNN presented in Section~\ref{sec:CNN} as a feature generator and then use these invariant features in a Linear Discriminant Analysis (LDA) classifier.

To do so, we removed the average pooling block and the fully connected layer used to train the system. Instead, we extracted $4$ local features (the average, standard deviation, maximum and minimum values) from the output of the spatial max pooling, \textit{i.e.} the feature maps containing the maximum activation to all filters within a rotation group in each pooling window. These $4\cdot M$ values, representing the distribution of the local responses of the filters within the image, are referred to as the \textit{local descriptors}. 

In order to add some information about the global frequency of the texture at a very low computational cost, we also stack to these features the mean, standard deviation, maximum and minimum values of the total cross power spectral density across orientations for each filter. The total cross power spectral density of two signals $x$ and $y$ is computed as:
\begin{equation}
	s = \sum_{ij}|\mathcal{F}(x)_{ij}\mathcal{F}(y)_{ij}|
\end{equation}
where $\mathcal{F}(x)_{ij}$ is the complex element in position $[i,j]$ of the Fourier transform of image $x$. Since we had already computed the Fourier transforms of both the images and the rotated filters to perform efficient convolutions, the additional cost of computing this terms is negligible. We refer to these $4\cdot M$ features as the \textit{global descriptors}.

The number of rotations for feature extraction was fixed to be $R=21$, even though the filters have been learned with $R=32$, since interpolating with a smaller $R$ resulted in a reduced computational load with no loss in accuracy 
Because we are using a shallow CNN that is being applied to the images at a single scale we use filters of $35\times 35$ pixels, a larger size than those typically used in the literature. This allows to capture texture information in a broad range of frequencies, enough to discriminate between the different texture classes.
Regarding the number of filters, we set it to $M=16$. Since we have $4\cdot M$ local descriptors and $4\cdot M$ global descriptors, we end up with $128$ features. These are reduced to $35$ by means of Principal Component Analysis (PCA) and used as features to train the LDA. When using only local or global features, PCA was applied to the single $64$ dimensional vectors.

\subsection{Standard data augmented CNN}

It could be argued that augmenting the training set with rotated versions of the original images and using them to train a standard CNN containing $M\cdot R$ filters would be equivalent to learning $M$ rotation groups of size $R$. In order to show that the features learned with the two approaches differ considerably, we trained the data-augmented CNN by randomly rotating input images in the range $[0,2\pi]$ at every training epoch.
This way, the same training sample is seen with different orientations during the training process. In order to avoid border effects, we cropped all the images used in this section, both with our approach and the standard one, to $88\times 88$ pixels after performing the rotation.

The training procedure was kept the same for both models. We trained without any weight decay until the cost function had dropped to half of its initial value. Then, it was trained for 100 more epochs with a weight decay of 0.1. Finally, the network was trained without weight decay until convergence. 
This procedure provides the same test accuracy than training the CNN with a fixed learning rate of 0.01, but with a shorter training.
We varied the total number of training samples from 1 to 20 per class. The learning rate was always set to $0.0001$ and the dropout rate to $0.2$. 


We chose $M=16$ and $R=32$ for the proposed system and compared it against 3 standard CNN models with $16$, $128$ and $512$ filters, respectively. We trained separate models for each one of these 4 settings, using training sets of varying size: from one single training image per class to the full set of 20 training images per class.


We report results obtained in the two configurations mentioned in Section~\ref{sec:CNN}: 1) The result given directly by the softmax classifier learned jointly with the filters by the CNN and 2) the full classification pipeline proposed in Section~\ref{sec:sota}, involving both local and global features. Note that the global features in the standard CNN can only provide one feature per filter: the total cross power spectral density between the image and the filter. Since we consider a small sample situation involving as little as a single training image per class, we cannot use the LDA classifier as in the comparisons against state-of-the-art (the covariance matrices become singular): we use a Nearest Neighbor ($k$NN) classifier with $k=1$ and using the cityblock distance instead.

\section{Results and Discussion}

\subsection{Comparison with state-of-the-art methods}

In Table~\ref{table:results} we show the classification results of our method along with several published results on the same benchmarks. Our method achieves accuracies comparable to the best published results~\cite{qi2015load} on the \textit{inca} and \textit{tl84} datasets and achieves the best results on the \textit{horizon} dataset.
Note how the simultaneous use the local and global descriptors seems to be required to reach state-of-the-art results, suggesting that they convey complementary information. The local features capture the abundance of a specific and local pattern, with independence on the orientation, due to the max-pooling across rotations. The global descriptor, on the other hand, captures the relationship between the frequencies occurring in the image and in the rotated versions of a filter.


\begin{table}[!t]
\caption{Classification accuracies on the three Outex benchmarks. Training always with inca illumination.}
\label{table:results}
\centering
\begin{tabular}{c||c|c|c}
\hline
Illumination in test:   & inca & tl84 & horizon\\
\hline\hline
Ojala \textit{et al.} (2002) \cite{ojala2002multiresolution} & 	97.9	& 90.2	& 87.2 \\\hline
Khellah (2011) \cite{khellah2011texture} & 	99.27	& 94.40	& 92.85 \\\hline
Zhao \textit{et al.} (2012) \cite{zhao2012completed} & 	99.38	& 94.98	& 95.51 \\\hline
Qi \textit{et al.} (2013) \cite{qi2013multi} & 		-	& 95.1	& - \\\hline
Liu \textit{et al.} (2012) \cite{liu2012extended} & 	99.7	& 98.7	& 98.1 \\\hline
Li \textit{et al.} (2013) \cite{li2013sensing} & 		99.17	& 98.91	& 98.22 \\\hline
Depeursinge \textit{et al.} (2014) \cite{depeursinge2014rotation} & 		98.4	& 97.8	& 98.4 \\\hline
Qi \textit{et al.} (2015) \cite{qi2015load} & 		\textbf{99.95}	& \textbf{99.65}	& 99.33 \\\hline
\hline
Ours (Local descriptors) & 						98.58	& 98.29	& 98.68 \\\hline
Ours (Global descriptors) & 						98.78	& 98.24	& 98.94 \\\hline
Ours (Local + global descriptors) & 						\textbf{99.95}	& 99.61	& \textbf{99.84} \\
\hline
\end{tabular}
\end{table}

\subsection{Comparison to data augmentation}

\begin{figure}[!t]
	\centering
	\begin{tabular}{c c}
		\includegraphics[width=.45\linewidth]{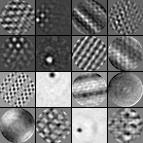} & \includegraphics[width=.45\linewidth]{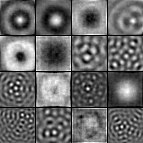}\\
		(a) & (b)\\
		\multicolumn{2}{c}{\includegraphics[width=.95\linewidth]{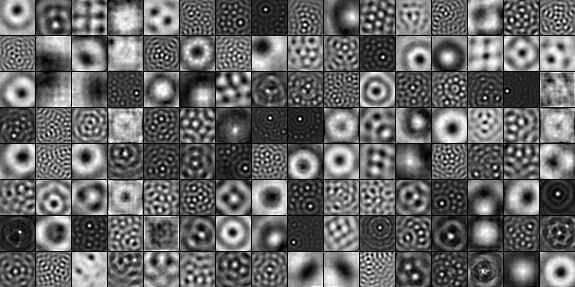}}\\
		\multicolumn{2}{c}{(c)}\\
	\end{tabular}
	\caption{(a) 16 rotatable $35\times 35$ filters learned by the proposed method showing clear directional patterns. 16 (b) and 128 (c) standard $35\times 35$ filters learned by the data-augmented CNNs show mostly non-directional patterns.}
	\label{fig:filters}
\end{figure}

In Fig.~\ref{fig:filters} we show a set of $M=16$ rotatable filters trained by our approach, with $R=32$ (Fig.~\ref{fig:filters}a), along with two sets of $M = 16$ and $M = 128$ filters learned by a standard CNN using randomized rotations of the input as data augmentation (Fig.~\ref{fig:filters}b and \ref{fig:filters}c). Note how the rotatable filters tend to learn highly directional patterns, some of them easily recognizable in the texture samples (Fig.~\ref{fig:data}), while the standard CNN learns non-directional filters representing different spatial frequencies and averages over orientations.\\

\begin{figure*}[ht]
	\centering
	\begin{tabular}{c c c}
		\multicolumn{3}{c}{\bf CNN softmax classification}\\
		\includegraphics[width=.28\linewidth]{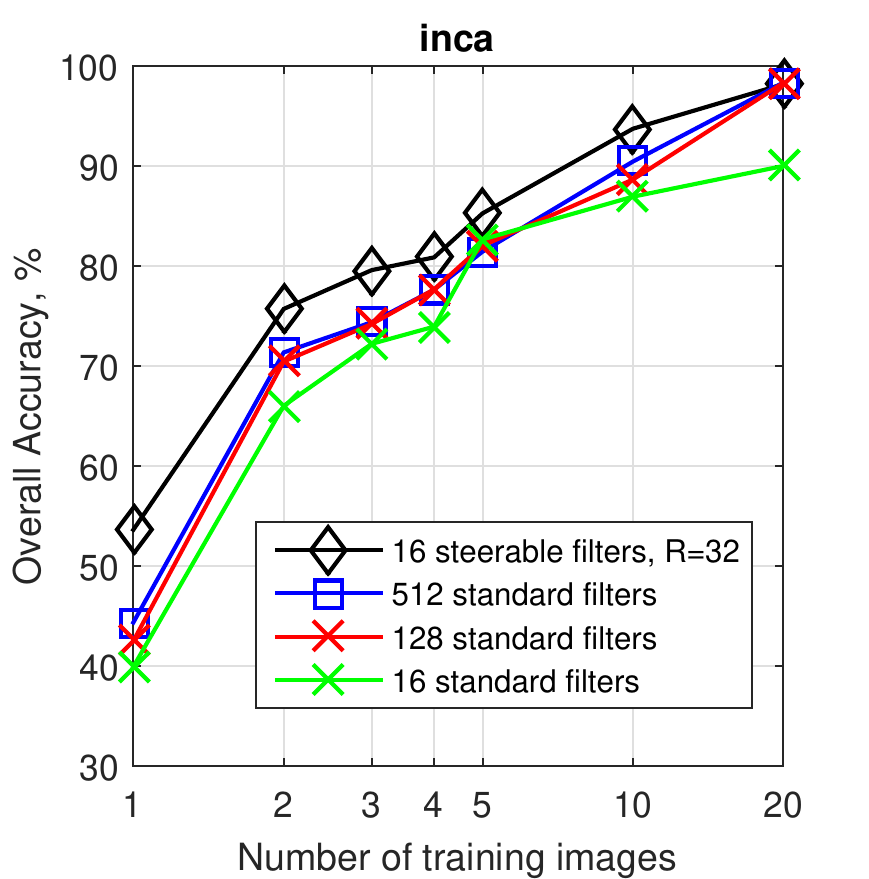} & \includegraphics[width=.28\linewidth]{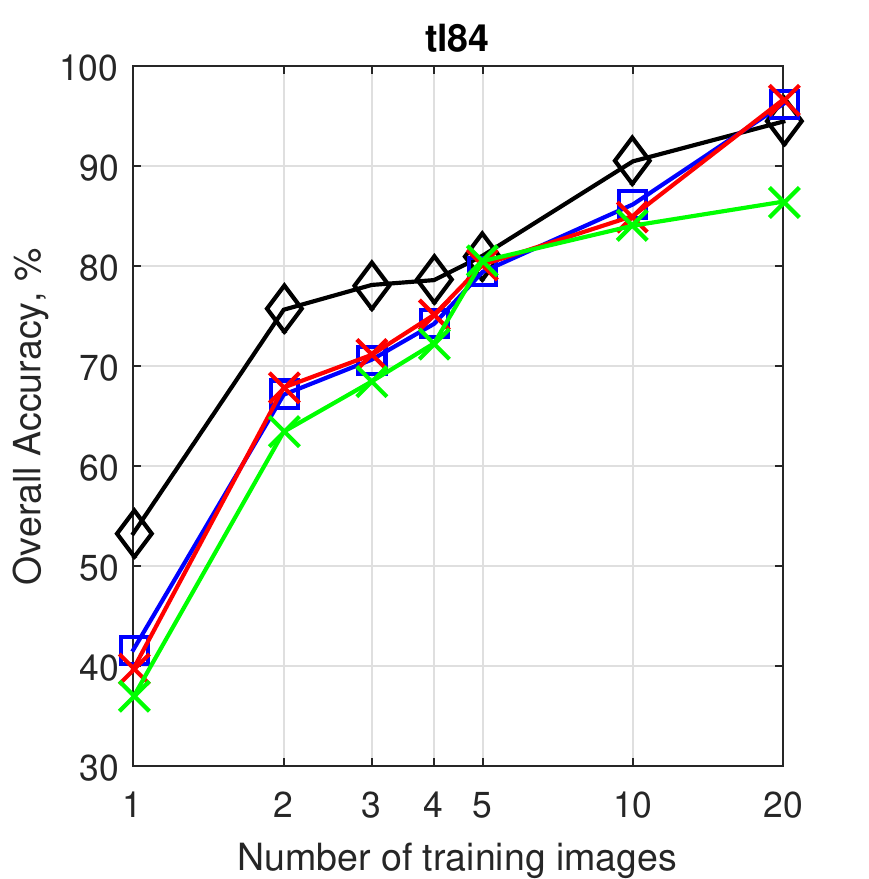} & 
		\includegraphics[width=.28\linewidth]{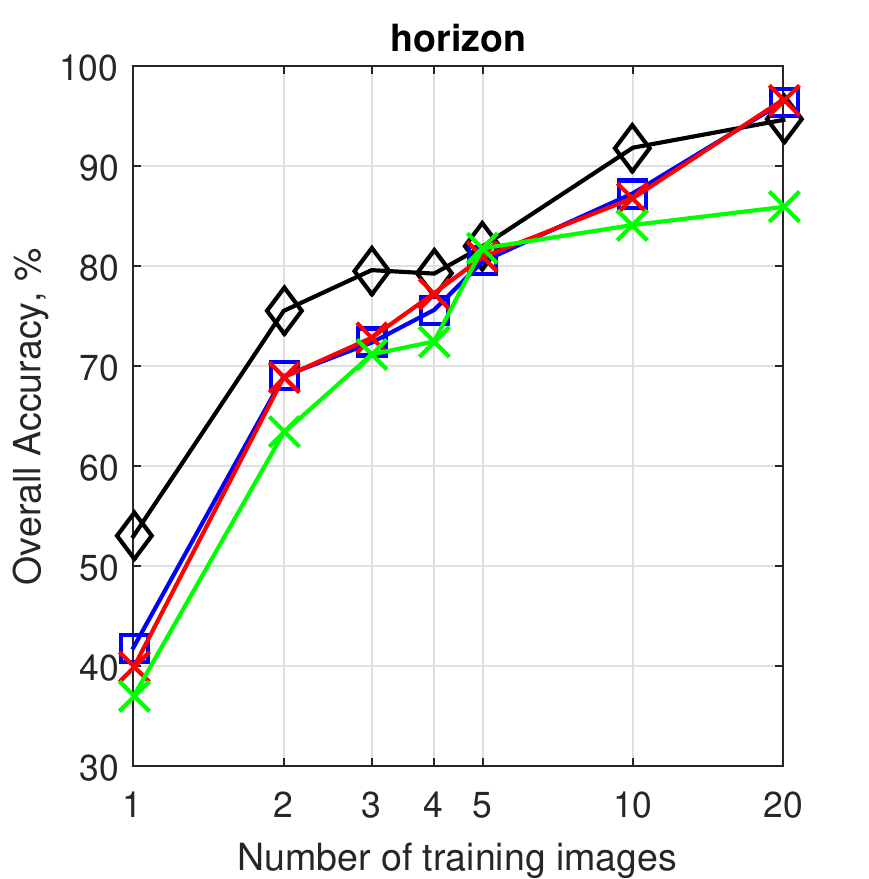}\\
		(a) & (b) & (c)\\
		{}\\
		\multicolumn{3}{c}{\bf Local and global descriptors with a $1$-NN classifier}\\
		\includegraphics[width=.28\linewidth]{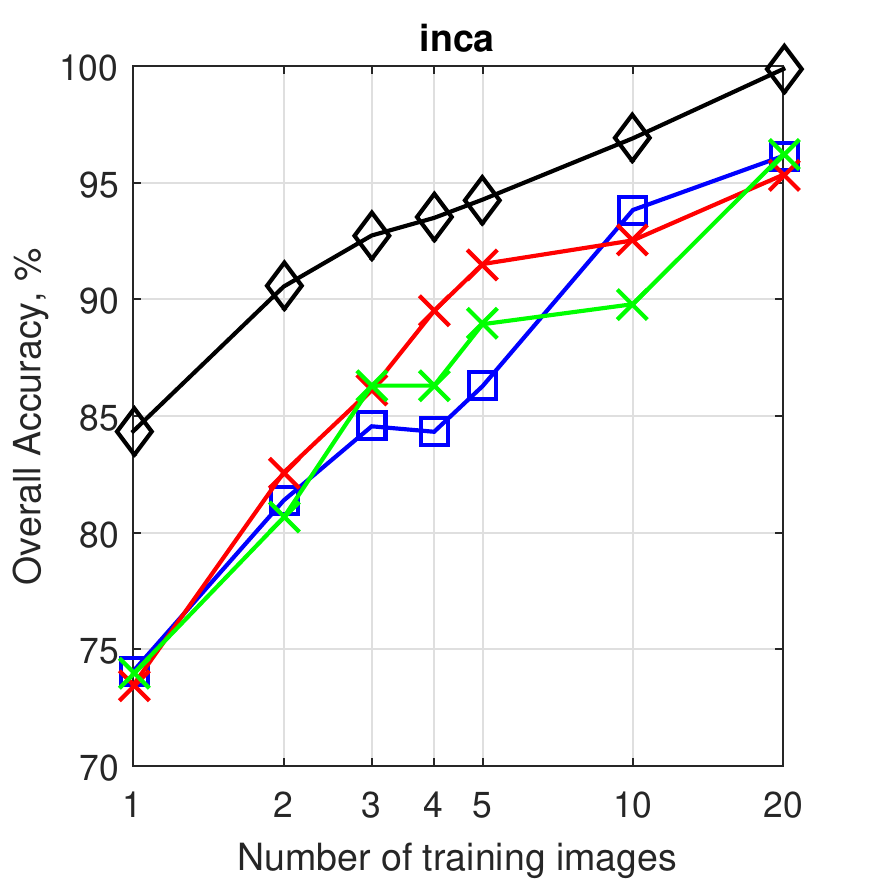} & \includegraphics[width=.28\linewidth]{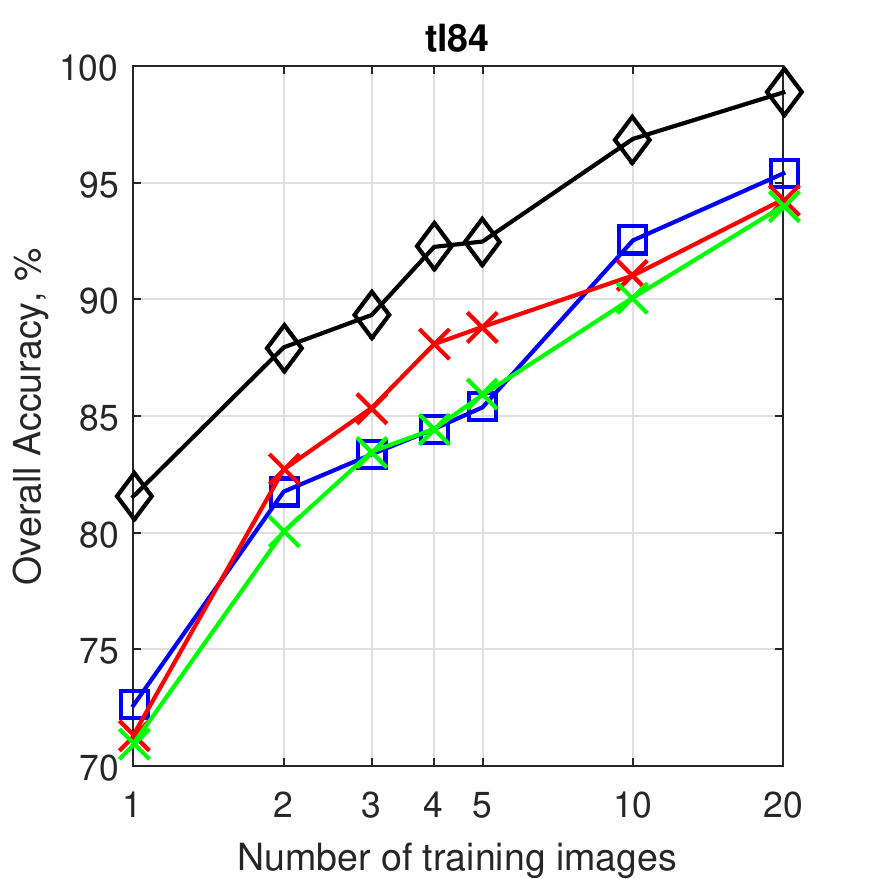} & 
		\includegraphics[width=.28\linewidth]{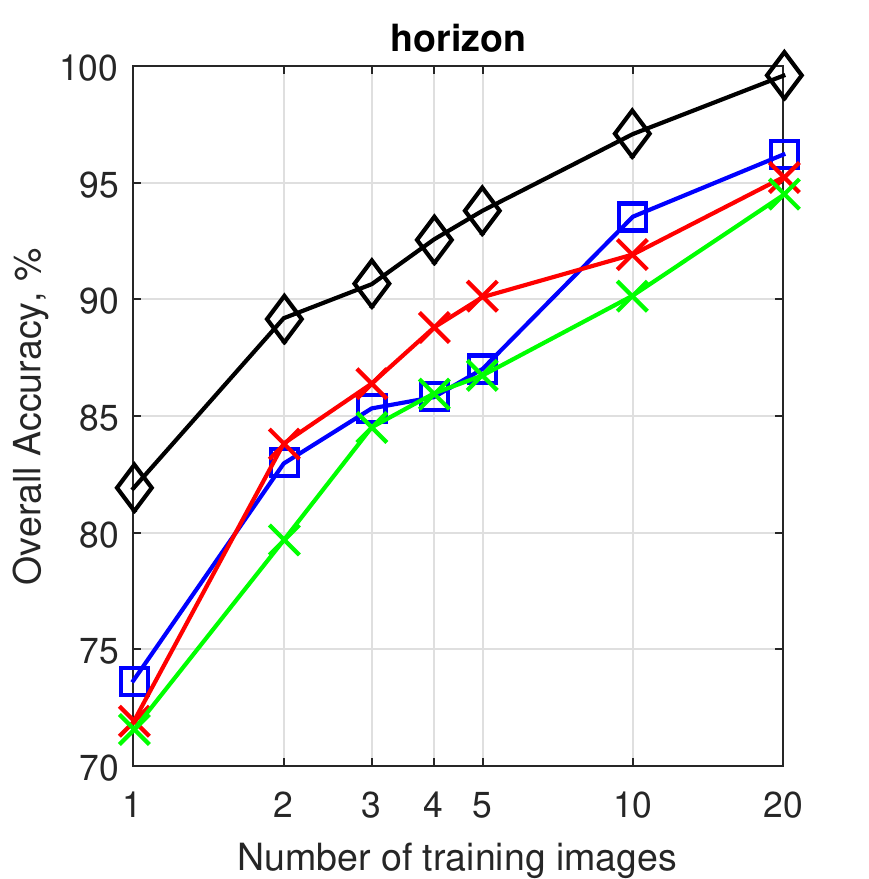}\\
		(d) & (e) & (f)\\
	\end{tabular}
	\caption{Results obtained by different CNNs on the 3 benchmarks using between 1 and 20 training images per class. Top row: results obtained by using the CNN directly as classifier (Fig.~\ref{fig:diag}). Bottom row: results using the extended feature space presented in Section~\ref{sec:sota} and the $1$-NN classifier. The advantage of using rotatable filters is larger when few training images are available.}
	\label{fig:train_n}
\end{figure*}

The results given by the softmax classifier learned by the CNN are shown in Figs.~\ref{fig:train_n}a, \ref{fig:train_n}b, \ref{fig:train_n}c for the 3 Outex benchmarks. The results obtained using the pipeline proposed in Section~\ref{sec:sota} are reported in Figs.~\ref{fig:train_n}d, \ref{fig:train_n}e, \ref{fig:train_n}f.

In both settings, we observe that the advantage of using rotatable filters is more pronounced when the number of training images becomes smaller. This is due to the smaller number of parameters to be learned (the same as in a standard CNN with $M=16$) and to the enhanced expressive power, due to the application of several rotated versions. This allows to learn informative and discriminative filters representing similar patterns that appear at different orientations within the images.  When considering the comparison with the standard data-augmented CNNs, the full set of 20 training images per class are required for the standard CNNs with $M=128$ and $M=512$ to perform better than the rotatable CNN with $M=16$. This means that standard data-augmented CNN start to profit from the larger number of learnable parameters only when the training set is sufficiently large and discriminative patterns are repeated often enough.

 In the second setting (the full classification pipeline), the results show the advantage of extending the output feature space beyond the average of local maximum activations. We can see how, after extending the feature space to better describe the distribution of the activations, the performance of the rotatable features becomes substantially higher than the one of the other standard CNN-based filterbanks. This effect can also be observed within the standard CNNs: with $M=16$, there is a boost in performance with 20 train images from an average of $87\%$ to $94\%$, while with $M=128$ and $M=512$ the performance is even reduced, from an average of $97\%$ to $95\%$. This similar performance among the 3 standard CNNs in the second setting suggests that it is not the higher number of filters that improves the performance of the raw CNN, but the higher number of mappings that can be learned in the fully connected layer. This helps explaining why using an extended feature space (as the one proposed in Section~\ref{sec:sota}) brings such a large performance improvement to the rotatable CNN.

\section{Conclusion}

In this paper, we proposed a strategy to learn explicitly rotation invariant rotatable filters by employing standard convolutional neural networks (CNN) formulations. We have shown the many advantages of explicitly accounting for rotation invariance when learning a discriminative filterbank for texture classification. We achieved these results by tying the weights of each group of filters in the first layer of a shallow CNN, such that each filter becomes a rotated version of the others in the group. The higher expressiveness of these rotatable filters and the subsequent reduction in the number of parameters to be learned provides an improvement in performance sufficient to meet the state-of-the art in a benchmark for rotation invariant texture classification. In addition, we have shown that the proposed methodology significantly outperforms standard data-augmented CNN, in particular in small training sets scenarios.

\vspace{.5cm}

\section*{Acknowledgment}
This work was supported in part by the Swiss National Science Foundation, via the
  grant 150593 ``Multimodal machine learning for remote sensing information fusion'' (http://p3.snf.ch/project-150593).



%



\bibliographystyle{IEEEtran}
\bibliography{cites}

\end{document}